\def\year{2021}\relax

\documentclass[letterpaper]{article} % DO NOT CHANGE THIS
\usepackage{aaai21}  % DO NOT CHANGE THIS
\usepackage{times}  % DO NOT CHANGE THIS
\usepackage{helvet} % DO NOT CHANGE THIS
\usepackage{courier}  % DO NOT CHANGE THIS
\usepackage[hyphens]{url}  % DO NOT CHANGE THIS
\usepackage{graphicx} % DO NOT CHANGE THIS
\usepackage{longtable}
\usepackage{multirow}
\usepackage{natbib}  % DO NOT CHANGE THIS AND DO NOT ADD ANY OPTIONS TO IT
\usepackage{caption} % DO NOT CHANGE THIS AND DO NOT ADD ANY OPTIONS TO IT
\usepackage{booktabs}
\usepackage{graphicx}
\usepackage{subcaption}
\usepackage[switch]{lineno}
\usepackage{amsmath}
\usepackage{xcolor}

\title{Batch Group Normalization}
\author{

    Xiao-Yun Zhou\textsuperscript{\rm 1}, Jiacheng Sun\textsuperscript{\rm 2}, Nanyang Ye\textsuperscript{\rm 3}, Xu Lan\textsuperscript{\rm 4}, Qijun Luo\textsuperscript{\rm 5}, Bo-Lin Lai \textsuperscript{\rm 3}, Pedro Esperanca\textsuperscript{\rm 2}, Guang-Zhong Yang \textsuperscript{\rm 3}, Zhenguo Li\textsuperscript{\rm 2}\\
}
\affiliations{
    \textsuperscript{\rm 1} Imperial College London, London, UK. email: xiaoyun.zhou27@gmail.com; \textsuperscript{\rm 2} Huawei Noah's Ark Lab; \textsuperscript{\rm 3} Shanghai Jiao Tong University, Shanghai, China; \textsuperscript{\rm 4} Queen Mary University of London, London, UK; \textsuperscript{\rm 5} Chinese University of Hong Kong, Shenzhen, China\\
}
\begin{document}
% \linenumbers

\maketitle

\begin{abstract}
Deep Convolutional Neural Networks (DCNNs) are hard and time-consuming to train. Normalization is one of the effective solutions. Among previous normalization methods, Batch Normalization (BN) performs well at medium and large batch sizes and is with good generalizability to multiple vision tasks, while its performance degrades significantly at small batch sizes. In this paper, we find that BN saturates at extreme large batch sizes, i.e., 128 images per worker\footnote{i.e., GPU}, as well and propose that the degradation/saturation of BN at small/extreme large batch sizes is caused by noisy/confused statistic calculation. Hence without adding new trainable parameters, using multiple-layer or multi-iteration information, or introducing extra computation, Batch Group Normalization (BGN) is proposed to solve the noisy/confused statistic calculation of BN at small/extreme large batch sizes with introducing the channel, height and width dimension to compensate. The group technique in Group Normalization (GN) is used and a hyper-parameter $\rm G$ is used to control the number of feature instances used for statistic calulation, hence to offer neither noisy nor confused statistic for different batch sizes. We empirically demonstrate that BGN consistently outperforms BN, Instance Normalization (IN), Layer Normalization (LN), GN, and Positional Normalization (PN), across a wide spectrum of vision tasks, including image classification, Neural Architecture Search (NAS), adversarial learning, Few Shot Learning (FSL) and Unsupervised Domain Adaptation (UDA), indicating its good performance, robust stability to batch size and wide generalizability. For example, for training ResNet-50 on ImageNet with a batch size of 2, BN achieves Top1 accuracy of $66.512\%$ while BGN achieves $76.096\%$ with notable improvement.
\end{abstract}

\section{Introduction}
\label{Sec:intro}
Since AlexNet was proposed in \cite{krizhevsky2012imagenet}, Deep Convolutional Neural Network (DCNN) has been a popular method for vision tasks including image classification \cite{deng2009imagenet}, object detection \cite{lin2014microsoft} and semantic segmentation \cite{pascal-voc-2012}.
DCNNs are usually composed of convolutional layers, normalization layers, activation layers, etc. Normalization layers are important in improving performance and speeding up training.

Batch Normalization (BN) was one of the early proposed normalization methods \cite{ioffe2015batch} and is widely used. It normalizes the feature map with the \textit{mean} and \textit{variance} calculated along with the batch, height, and width dimension of a feature map and then re-scales and re-shifts the normalized feature map to ensure DCNN representation ability. Based on BN, many normalization methods for other tasks have been proposed. For example, Layer Normalization (LN) was proposed for calculating the statistics along the channel, height and width dimension for Recurrent Neural Network (RNN) \cite{ba2016layer}. Weight Normalization (WN) was proposed to parameterize the weight vector for supervised image recognition, generative modelling, and deep reinforcement learning \cite{salimans2016weight}. Divisive Normalization which includes BN and LN as special cases was proposed for image classification, language modeling and super-resolution \cite{ren2016normalizing}. Instance Normalization (IN) where the statistics were calculated from the height and width dimension was proposed for fast stylization \cite{ulyanov2016instance}. Instead of calculating the statistics from data, Normalization Propagation estimated them data-independently from the distribution in layers \cite{arpit2016normalization}. Group Normalization divided the channels into groups and calculated the statistics for each grouped channel, height and width dimension, showing stability to batch sizes \cite{wu2018group}. Positional Normalization (PN) was proposed to calculate the statistics along the channel dimension for generative networks \cite{li2019positional}.

Among these normalization methods, BN can usually achieve good performance at medium and large batch sizes. However, its performance degrades at small batch sizes, as shown in pre works \cite{wu2018group,ioffe2017batch}. Furthermore, as shown in our experiments, BN's performance saturates at extreme large batch sizes, i.e., 128 images per worker. GN enjoys a greater degree of stability at different batch sizes, while slightly under-performs BN at medium and large batch sizes. Other normalization methods, including IN, LN and PN perform well in specific tasks, but are usually less generalizable to and under-perform in other vision tasks. As reviewed in Related Work, many works have been conducted on proposing new normalization methods with good performance, stability and generalizability.

In this paper, unlike those reviewed works where additional trainable parameters, extra computation or additional information are used, Batch Group Normalization (BGN) is parameter- and computation-efficient. We know the fact that mini-batch training usually can perform better than single-batch (use a single image example as the DCNN input per iteration) and all-batch training (use all image examples as the DCNN input per iteration), as single-batch training can indicate noisy gradient while all-batch training may not indicate representative gradient (each image example indicates gradient with different directions, thus, adding them all indicates confused gradient). Inspired by this fact, we think the number of feature instances in statistic calculation in normalization should also be moderate, i.e., the degraded/saturated performance of BN on small/extreme large batch sizes is due to noisy/confused statistic calculation. 

\begin{figure}[!t]
    \begin{subfigure}[t]{1.0\columnwidth}
        %\centering
        \includegraphics[width=\textwidth]{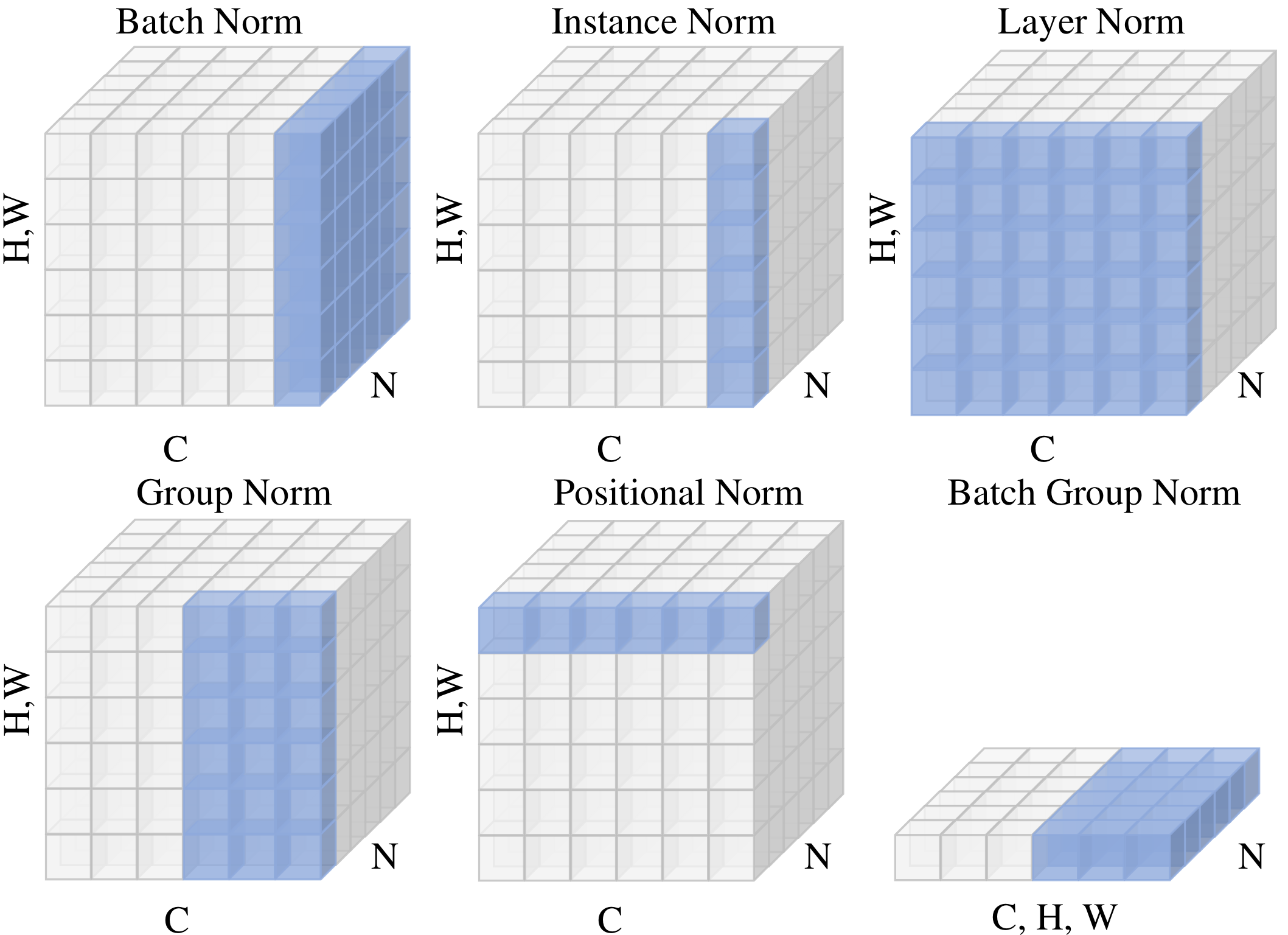}
    \end{subfigure}
    \begin{subfigure}[t]{1.0\columnwidth}
        %\centering
        \includegraphics[width=\textwidth]{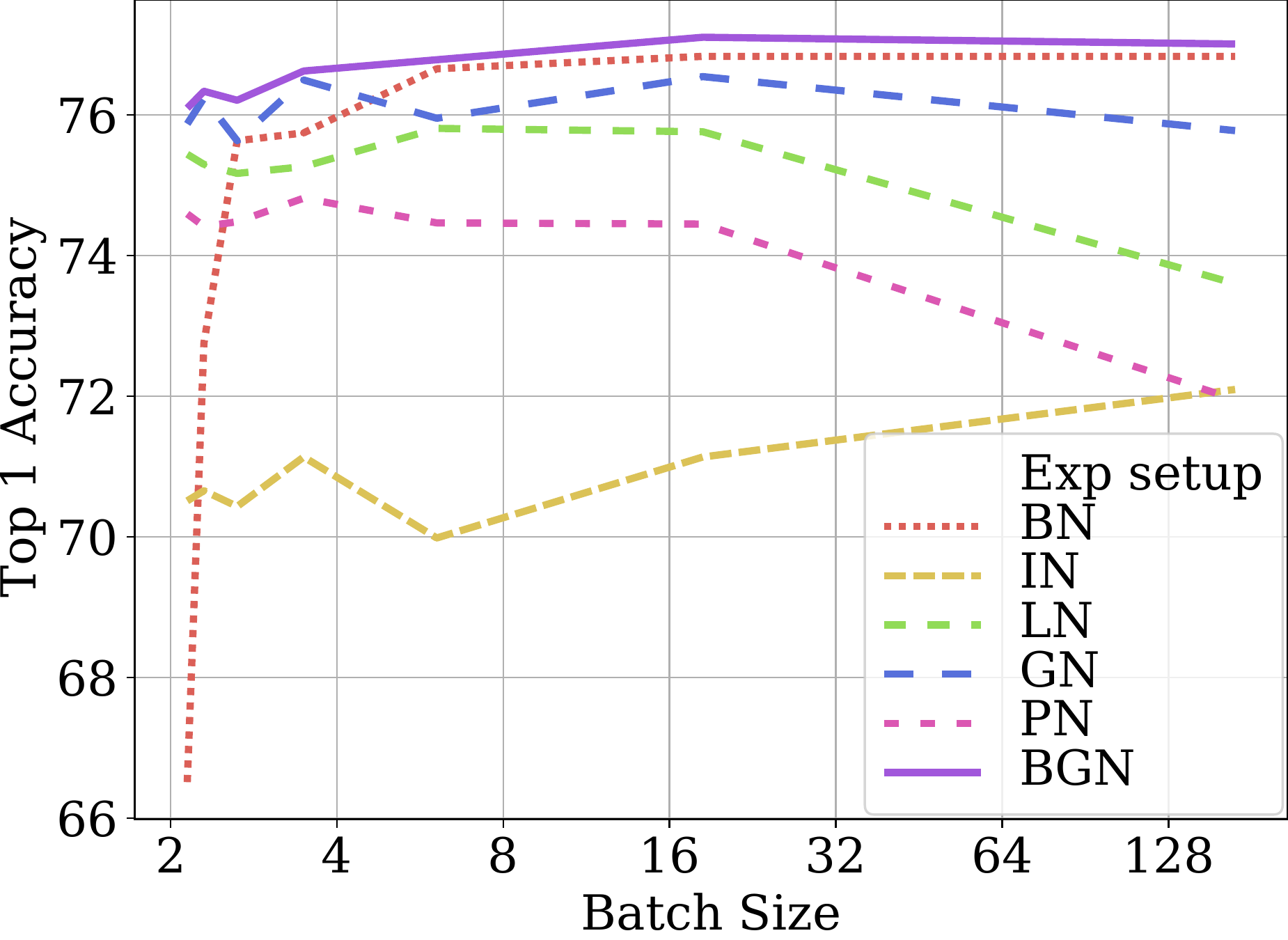}
    \end{subfigure}
\caption{top). The difference between the statistic calculation in BN, IN, LN, GN, PN and BGN. Each subplot shows a feature map tensor, with N, C and (H, W) as the batch, channel and spatial axes. The pixels in blue are used to compute the statistics. This figure is inspired by \cite{wu2018group}. bottom). The Top1 accuracy of training ResNet-50 on ImageNet, with different batch sizes, and with BN, IN, LN, GN, PN and BGN as the normalization. }
\label{Fig:demo_normalization}
\end{figure}

Hence, BGN is proposed to facilitate the performance degradation/saturation of BN at small/extreme large batch sizes with the group technique from GN. It merges the channel, height and width dimensions into a new dimension, divides the new dimension into feature groups, and calculates the statistics across the whole mini-batch and feature group. A hyper-parameter $\rm G$ is used to control the level of feature division and to supply proper statistics for different batch sizes. The difference between BN, IN, LN, GN, PN and BGN, with respect to the dimensions along which the statistics are computed, are illustrated in Fig.~\ref{Fig:demo_normalization}(top). The Top1 accuracy of training ResNet-50 on ImageNet by using BN, IN, LN, GN, PN and BGN as the normalization layer are shown in Fig.~\ref{Fig:demo_normalization}(bottom). Without adding trainable parameters, using extra information or requiring extra computation, BGN achieves both good performance and stability at different batch sizes. We test on Neural Architecture Search (NAS), adversarial learning, Few Show Learning (FSL) and Unsupervised Domain Adaptation (UDA), BGN outperforms BN, IN, LN, GN and PN, showing its good generalizability.

% \begin{figure}[!t]
%     \begin{subfigure}[t]{0.3\columnwidth}
%         %\centering
%         \includegraphics[width=\textwidth]{accuracy_ImageNet_train_BN.pdf}
%         \caption{BN train}
%         \label{fig:ImageNet_train_BN}
%     \end{subfigure}
%     \begin{subfigure}[t]{0.3\columnwidth}
%         %\centering
%         \includegraphics[width=\textwidth]{accuracy_ImageNet_train_GN.pdf}
%         \caption{GN train}
%         \label{fig:ImageNet_train_GN}
%     \end{subfigure}
%     \begin{subfigure}[t]{0.3\columnwidth}
%         %\centering
%         \includegraphics[width=\textwidth]{accuracy_ImageNet_train_BGN.pdf}
%         \caption{BGN train}
%         \label{fig:ImageNet_train_BGN}
%     \end{subfigure}
% \caption{The training accuracy of BN, GN and the proposed BGN at different batch sizes of 128, 64, 32, 16, 8 and 4 at each epoch, with training ResNet-50 on ImageNet for image classification.}
% \label{Fig:convergence_batchsize}
% \end{figure}

\section{Related Work}
\label{Sec:work}

\textbf{Why normalization works?} The effectiveness of BN has been attributed to \textit{internal covariate shift} where the distribution of each layer's input changes and hence lower learning rate and careful parameter initialization are essential to guarantee good training in DCNNs without normalization \cite{ioffe2015batch}.
Other works have investigated the reasons for the success of normalization.
For example, Santurkar \textit{et al.} \cite{santurkar2018does} proposed that the effectiveness of BN has little to do with \textit{internal covariate shift}, but to make the optimization landscape smoother and hence to introduce stable gradients and faster training.
Bjorck \textit{et al.} \cite{bjorck2018understanding} proposed that allowing larger learning rates is the main reason for BN to achieve faster convergence and better generalization.
A theoretical support was supplied for the effectiveness of BN in tuning learning rates less \cite{arora2019theoretical}.
Luo \textit{et al.} demonstrated that BN and regularization share the same traits \cite{luo2018towards}.
A quantitative analysis was provided to compare the gradient descent with and without BN on Ordinary Least Squares (OLS) \cite{cai2019quantitative}.
In \cite{fan2017revisit}, it was shown with fuzzy neural networks that BN estimates the correct bias induced by the generalized hamming distance.
In contrast, BN was proved to be the cause of the gradient explosion for DCNNs without residual learning \cite{yang2018a}.
In \cite{li2019understanding}, from the theoretical and statistical aspects, the disharmony between dropout and BN was explained.
It was claimed that BN is not unique for stable training, higher learning rates, accelerated convergence and improved generalization, and can be replaced by better initialization \cite{zhang2018residual,de2020batch}. Normalization layers were shown to introduce a stable gradient magnitude when training the Long-Short-Term Memory (LSTM) \cite{hou2019normalization}. Even though many interesting theories have been proposed to explain the effectiveness of BN, there is still a lack of consensus.

\textbf{Improvements} Previous works can be further improved. Centered WN proposed to add a learnable parameter to adjust the weight norm in WN \cite{huang2017centered}. Recurrent BN was proposed to not only apply BN in the input-to-hidden, but also to the hidden-to-hidden transformation of RNNs \cite{cooijmans2016recurrent}. Batch Renormalization was proposed to decrease the dependence of BN on the size of mini-batches \cite{ioffe2017batch}. Riemannian approach was combined to BN on the space of weight vectors, which improves BN's performance on various DCNN architectures and datasets \cite{cho2017riemannian}. EvalNorm was proposed to estimate the corrected normalization statistics during evaluation to address the performance degradation of BN \cite{singh2019evalnorm}. Moving Average BN proposed using the batch statistics in the backward propagation of traditional BN \cite{yan2019towards}. Based on LN, Adaptive Normalization proposed to modify bias and gain by using a new transformation function \cite{xu2019understanding}. Root Mean Square LN proposed to abandon the re-centering and keep the re-scaling in LN \cite{zhang2019root}.

Researchers are also looking into non-linear normalizing techniques with extra computation. For example, ZCA was used to replace the centering and scaling calculations in BN, resulting in Decorrelated BN \cite{huang2018decorrelated}. Iterative Normalization proposed using Newton iterations to avoid the eigen-decomposition in Decorrelated BN, indicating much more efficient whitening \cite{huang2019iterative}. Instead of normalizing in the spatial space, Spectral Normalization was proposed to normalize the spectral norm of weights and was used in Generative Adversarial Networks (GANs) \cite{miyato2018spectral} and adversarial training \cite{farnia2018generalizable}.

Learning-to-normalize is also explored. Different normalization methods introduced before can be combined to achieve better performance and generalization. Batch-Instance Normalization (BIN) was proposed for adaptively style-invariant neural networks, with using a trainable parameter to combine the feature maps calculated from BN and IN \cite{nam2018batch}. Switchable Normalization (SN) was proposed by using trainable parameters on the \textit{mean} and \textit{variance} calculated in BN, IN and LN to learn new statistics \cite{luo2018differentiable,luo2019switchable}. Sparse SN used SparseMax to make the trainable parameters in SN sparse \cite{shao2019ssn}. In Instance-Level Meta Normalization \cite{jia2019instance}, feature feed-forward and gradient back-propagation were used to learn the normalization parameters.

\textbf{Others} Normalization can be used as a method to achieve a task directly. For example, a real-time and arbitrary style transfer was achieved by aligning the \textit{mean} and \textit{variance} of the content features to that of the style features \cite{huang2017arbitrary}. Domain adaptation was achieved by changing the statistics from the source domain to the target domain \cite{li2018adaptive}.
New tasks can be solved by using BN to combine learned tasks \cite{data2018interpolating}.
Specific normalization methods have also been specifically proposed for cross-domain tasks \cite{li2019efficient,wang2019transferable}, global covariance pooling networks \cite{li2018towards}, multitask learning \cite{chen2018gradnorm,deecke2018mode}, UDA \cite{chang2019domain}, semantic image synthesis \cite{park2019semantic}, medical area \cite{zhou2019normalization,zhou2019u,zhou20193d}, and scene text detection \cite{xu2019geometry}.
Kalman Normalization (KN) was proposed to combine internal representations across multiple DCNN layers \cite{wang2018kalman}. And, instead of using $L^2$ norm, $L^1$ and $L^{\infty}$ norm were proposed for numerical stability in low-precision calculation \cite{hoffer2018norm}.

\section{Methodology}
\label{Sec: method}

In a DCNN with $\rm L$ layers, for an input feature map $\mathcal{F}^l, l\in[1, \rm L]$, which is usually with four dimensions $(\rm N, C, H, W)$, where $\rm N, C, H, W$ is the batch, channel, height and width dimension respectively. For simplification, $n \in \rm [1,N], c \in \rm [1,C], h \in \rm [1,H], w \in \rm [1,W]$ are the corresponding batch, channel, height and width indices and will not be repeatly defined in the following usage. The feature map at $(l+1)$th layer is calculated as:
\begin{equation}
    \mathcal{F}^{l+1} = \psi (\theta (\phi (\mathcal{F}^l, w^l, b^l), \gamma^l, \beta^l) )
\end{equation}
where $w^l$ and $b^l$ are the trainable weight and bias parameters in convolutional layers, $\gamma^l$ and $\beta^l$ are the trainable re-scale and re-shift parameters in normalization layers, $\psi (\cdot)$ is the activation function. $\theta (\cdot)$ is the normalization function. $\phi (\cdot)$ is the convolusion function.

A typical normalization layer includes four steps: (1) divide the feature map into feature groups; (2) calculate the $mean$ and $variance$ statistics for each feature group; (3) normalize each feature group with the calculated statistics; (4) re-scale and re-shift the normalized feature map to maintain the DCNN representation ability. For example, in \textbf{BN}, the feature map is divided along the channel dimension, the $mean$ $\mu_c$ and $variance$ $\delta_c^2$ are calculated along the batch, height and width dimension as:
\begin{equation}
\label{Equ:mean}
\mu_c = \frac{1}{\rm N \times \rm H \times \rm W} \sum\limits_{n=1}^{\rm N} \sum\limits_{h=1}^{\rm H} \sum\limits_{w=1}^{\rm W} f_{n,c,h,w}
\end{equation}

\begin{equation}
\label{Equ:var}
\delta_c^2 = \frac{1}{\rm N \times \rm H \times \rm W} \sum\limits_{n=1}^{\rm N} \sum\limits_{h=1}^{\rm H} \sum\limits_{w=1}^{\rm W} (f_{n,c,h,w}-\mu_c)^2
\end{equation}

Then the feature map is normalized as:
\begin{equation}
\begin{aligned}
\hat{f}_{n,c,h,w} = \frac{f_{n,c,h,w}-\mu_c}{\sqrt{\delta_c^2+\epsilon}} 
\end{aligned}
\end{equation}
$\epsilon$ is a small number added for division stability. In order to maintain the DCNN representation ability, extra trainable parameters are added for each feature channel:

\begin{equation}
\label{Equ:gamma}
f'_{n,c,h,w} = \gamma_c \hat{f}_{n,c,h,w}+\beta_c 
\end{equation}

\begin{table*}[h]	
	\setlength{\tabcolsep}{0.07cm} 
	\caption{The Top1 validation accuracy of BGN on ImageNet with ResNet-50. G is set to be from 512 to 1.}
	\centering
	\begin{tabular}{c|cccccccccc}
	    \hline %\toprule
    	\multirow{2}{*}{Batch size} & \multicolumn{10}{c}{Group number}\\
		\cline{2-11}
		& 512 & 256 & 128 & 64 & 32 & 16 & 8 & 4 & 2 & 1\\
		\hline
	    128 & \textbf{77.008} & 75.936	& 76.256 & 76.096 & 75.984 & 75.504 & 76.016 & 75.632 & 75.520 & 75.360 \\
		\hline
		2 & 63.968 & 66.000 & 68.160 & 70.064 & 71.856 & 72.832 & 74.928 & 75.120 & 75.968 & \textbf{76.096}
\\
		\hline
	\end{tabular}
\label{Tab:ImageNet-G}
\end{table*}

By including multiple examples into statistic calculation, BN enjoys good performance at medium and large batch sizes and good generalizability to multiple vision tasks, i.e. NAS. However, its performance degrades dramatically, i.e., $>10\%$ in our ImageNet experiment, at small batch sizes. In order to improve this shortage, \textbf{GN} includes grouped channel dimension into statistic calculation:
\begin{equation}
\mu_{n,g} = \frac{1}{\rm M \times \rm H \times \rm W} \sum\limits_{m=(g-1)\cdot \rm M+1}^{\rm g\cdot M} \sum\limits_{h=1}^{\rm H} \sum\limits_{w=1}^{\rm W} f_{n,m,h,w}
\end{equation}

\begin{equation}
\delta_{n,g}^2 = \frac{1}{\rm M \times \rm H \times \rm W} \sum\limits_{m=(g-1)\cdot \rm M+1}^{\rm g\cdot M} \sum\limits_{h=1}^{\rm H} \sum\limits_{w=1}^{\rm W} (f_{n,m,h,w}-\mu_{n,g})^2
\end{equation}

where $g \in \rm [1,G]$, $\rm G$ is a hyper parameter - group number, $\rm M=C//G$, $//$ is floor division. GN enjoys good stability to different batch sizes, however, its performance is slightly lower than BN at medium and large batch sizes and its generalizability to other vision tasks is weaker than BN. Except these known phenomena, our experiments show that BN saturates at extreme large batch sizes. 

We think the degradation/saturation of BN at small/extreme large batch sizes are caused by noisy/confused statistic calculation. Similar indication also exists in mini-batch training, where single-batch/all-batch training are usually worse than mini-batch training, as noisy/confused gradients are calculated. To facilitate this, we propose \textbf{BGN} where the number of feature instances used for statistic calculation is controlled to be proper by using the group technique in GN. To be in more details, we first merge the channel, height and width dimensions into a new dimension and achieve $\rm \mathcal{F}^l_{N \times D}$, where $\rm D = C \times H \times W$. The mean $\mu_g$ and variance $\delta_g^2$ are calculated along the batch and new dimension as:
\begin{equation}
\mu_g = \frac{1}{\rm N \times \rm S}  \sum\limits_{n=1}^{\rm N} \sum\limits_{d=(g-1)\cdot \rm S+1}^{\rm g\cdot S} f_{n,d}
\end{equation}

\begin{equation}
\delta_g^2 = \frac{1}{\rm N \times \rm S}  \sum\limits_{n=1}^{\rm N} \sum\limits_{d=(g-1)\cdot \rm S+1}^{\rm g\cdot S} (f_{n,d}-\mu_g)^2
\end{equation}

where $\rm G$ is the number of groups that the new dimension is divided and is a hyper-parameter, $\rm S=M/G$ is the number of instances inside each divided feature group. When the batch size is small, a small G is used to combine the whole new dimension into statistic calculation to avoid noisy statistics, while when the batch size is large, a large G is used to split the new dimension into small pieces for calculating the statistics to avoid confused statistics. $\gamma_c$ and $\beta_c$ are used in the same way as BN. In BN, $\mu_c$ and $\delta_c^2$ used in the test are the moving average of that in the training stage. The proposed BGN uses this policy as well.

\textbf{Relation to General Batch Group Normalization (GBGN)} GBGN \cite{summers2019four} could be confused as a similar work to BGN, we clarify our contributions as below:
1. We first propose the saturation of BN at extreme large batch sizes.
2. We first propose that the degradation/saturation of BN at small/extreme large batch size are caused by noisy/confused statistic calculation.
3. We propose to use group in the channel, height and width dimension to compensate (in GBGN, the group is along the batch and channel dimension).
4. We offer extensive experiments on image classification, NAS, adversarial learning, FSL and UDA to validate our thoughts.

\section{Experiments}

Among the reviewed normalization methods, BN, IN, LN, GN, and PN\footnote{Re-injection path was used in the original PN \cite{li2019positional}, in this paper, for a fair comparison, it is not included.} are suitable for being baselines of BGN, as other normalization methods usually use additional trainable parameters, non-linear normalization or information from multiple layers, which are orthogonal to and can be combined to BGN to improve its performance further. BGN is validated on applications, including image classification, NAS, adversarial training, FSL and UDA. 

\subsection{Image Classification on ImageNet with ResNet-50}
\label{sec:image classification}
Image classification is one of the applications used to validate BGN. We focus on ImageNet \cite{krizhevsky2012imagenet} which contains $\rm 1.28M$ training images and $50000$ validation images. The model used is ResNet-50 \cite{he2016deep}.

\textbf{Implementation details:} 8 GPUs are used in all ImageNet experiments. The gradients used for backpropagation are averaged across 8 GPUs, while the \textit{mean} and \textit{variance} used in BN and BGN are calculated within each GPU. $\gamma_c$ and $\beta_c$ are initialized as 1 and 0 respectively, while all other trainable parameters are initialized as in \cite{he2016deep}. 120 epochs are trained with the learning rate decayed by $10\times$ at the 30th, 60th, and 90th epoch. The initial learning rates for the experiments with batch sizes of 128, 64, 32, 16, 8, 4 and 2 are 0.4, 0.2, 0.1, 0.05, 0.025, 0.0125 and 0.00625 respectively, following \cite{goyal2017accurate}. Stochastic Gradient Descent (SGD) is used as the optimizer. A weight decay of $10^{-4}$ is applied to all trainable parameters. For the validation, each image is cropped into $224\times224$ patches from the center, and the Top1 accuracy is reported. Following \cite{wu2018group}, the median accuracy in the last five epochs is reported to reduce the random variance. All experiments are trained under the same programming implementation, with replacing the normalization layer into BN, IN, LN, GN, PN, and BGN respectively.

\textbf{Hyper-parameter - $\rm G$:} to explore the hyper-parameter $\rm G$, BGN with a group number of 512, 256, 128, 64, 32, 16, 8, 4, 2 and 1 respectively are used as the normalization layer in ResNet-50 for ImageNet classification. The largest (according to GPU memory) and smallest batch size in our experiments - 128 and 2 are tested. The Top1 accuracy of the validation dataset is shown in Tab. \ref{Tab:ImageNet-G}. We can see that a large G - 512 is suitable for a large batch size - 128, while a small G - 1 is suitable for a small batch size - 2. It can support our claims that a proper number of feature instances is important for the statistic calculation in normalization. When the batch size is large/small, a large/small $\rm G$ is used to split/combine the new dimension to avoid confused/noisy statistic calculation. The accuracy variance for batch size 128 is smaller than that for batch size of 2, indicating that either saturation is less serious than degradation in normalization or a large batch size of 128 has not reached the saturation edge yet.
% 512 is the largest batch size our GPU can hold, the accuracy variance may be larger if a larger batch size can be tested.

\textbf{Comparison with baselines:}  BN, IN, LN, GN, PN, GBGN and BGN are used as the normalization layer in ResNet-50, with a batch size of 128, 64, 32, 16, 8, 4 and 2 respectively. The group number in GN is set as 32, which was claimed as the optimal configuration for GN \cite{wu2018group}. The group number for channel is set as 32 while that for batch is set as the batch size for GBGN based on our experience. With this setting, GBGN equals to BGN at moderate batch sizes, hence GBGN is only included in ImageNet experiments and is ignored in the following few experiments where moderate batch sizes are used. The group number in BGN is set to be 512, 256, 128, 64, 16, 2 and 1 for batch sizes of 128, 64, 32, 16, 8, 4 and 2 respectively. We choose $\rm G$ for the largest and smallest batch size according to Tab. \ref{Tab:ImageNet-G} while choose $\rm G$ for other batch sizes with interpolation. The Top1 accuracy is shown in Tab. \ref{Tab:ImageNet-batchsize}. We can see that the proposed BGN out-performs all previous methods, including BN, IN, LN, GN, PN and GBGN at all different batch sizes. To be specific, BN approaches BGN's performance at large batch sizes, however, its performance degrades quickly at small batch sizes. GBGN is proposed for small batch sizes, however it under-performs BGN with $4.24\%$ at the batch size of 2, indicating the importance of introducing the whole channel, height and width dimension to compensate the noisy statistic calculation. IN overall performs not well on ImageNet classification. LN, GN and PN achieve average Top1 accuracy of $75.191\%$, $76.073\%$, $74.167\%$ respectively, while the proposed BGN achieves higher average Top1 accuracy of $76.594\%$.

\begin{table}[h]	
	\setlength{\tabcolsep}{0.07cm} 
	\caption{The Top1 validation accuracy of BN, IN, LN, GN, PN, GBGN and BGN on ImageNet with ResNet-50 and with different batch sizes from 128 to 2.}
	\centering
	\begin{tabular}{c|ccccccc}
		\hline
		\multirow{2}{*}{method} & \multicolumn{6}{c}{Batch size}\\
		\cline{2-8}
		& 128 & 64 & 32 & 16 & 8 & 4 & 2\\
		\hline
	    BN & 76.832 & 76.832 & 76.656 & 75.744 & 75.632 & 72.752 & 66.512\\
	    \hline
	    IN & 72.096 & 71.136 & 69.984 & 71.136 & 70.432 & 70.656 & 70.512\\
	    \hline
	    LN & 73.600 & 75.760 & 75.808 & 75.264 & 75.168 & 75.296 & 75.440\\
		\hline
		GN & 75.776 & 76.544 & 75.952 & 76.496 & 75.632 & 76.240 & 75.872\\
		\hline
	    PN & 71.952 & 74.448 & 74.464 & 74.816 & 74.48 & 74.416 & 74.592\\
		\hline
		GBGN & 75.984 & 76.624 & 76.368 & 76.592 & 75.840 & 75.024 & 71.856\\
		\hline
	    BGN & \textbf{77.008} & \textbf{77.104} & \textbf{76.784} & \textbf{76.624} & \textbf{76.208} & \textbf{76.336} & \textbf{76.096}\\
		\hline
	\end{tabular}
\label{Tab:ImageNet-batchsize}
\end{table}

\subsection{Image Classification on CIFAR-10 with NAS}
\label{sec:nas}

Except manually designed and regular DCNN, BGN is applicable to automatically-designed and less-regular ones as well. We experiment with cell-based architectures designed automatically with NAS, specifically DARTS \cite{liu2018darts} and Multi-agent Neural Architecture Search (MANAS \cite{MANAS2019}). For DARTS, we experiment with normalization methods for both the searching and training. For MANAS, we experiment with normalization methods for the training only.
% In MANAS and other NAS methods sharing the same search space, the family of architectures searched (the \textit{search space}; see Fig. \ref{fig:cells}) is composed of a sequence of cells, where each cell is a directed acyclic graph with nodes representing feature maps and edges representing network operations, e.g. convolutions or pooling layers (\cite{MANAS2019} and references therein). Given a set of possible operations, MANAS uses a multi-agent learning approach (the \textit{search strategy}) to find the combination of operations leading to the best-performing architecture according the validation accuracy.

\begin{figure}[!t]
\includegraphics[width=0.48\textwidth]{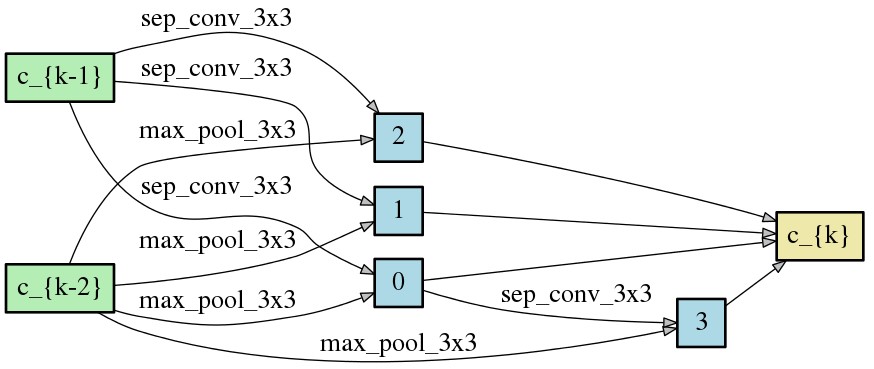}%
\hspace{0.02\textwidth}%
\includegraphics[width=0.48\textwidth]{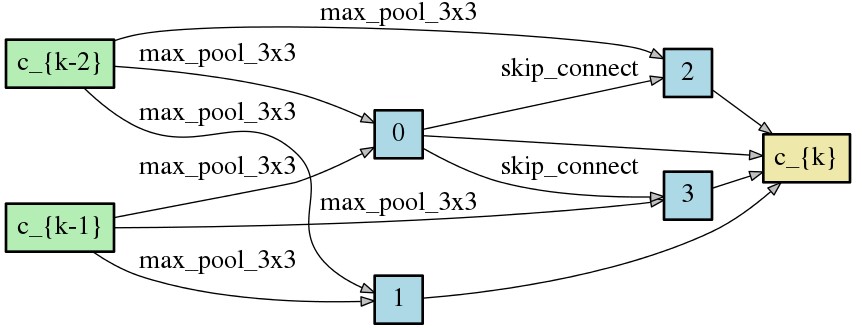}%
\caption{Cell-based architectures. Example of a \textit{normal cell} (top) and a \textit{reduction cell} (bottom) in the DARTS and MANAS search space. Each cell has $2$ input (green), $4$ internal (blue) and $1$ output node (yellow). Multiple cells are connected in a feedforward fashion to create a DCNN.
}
\label{fig:cells}
\end{figure}

DARTS and MANAS share the same search space, the family of architectures searched (the \textit{search space}; see Fig. \ref{fig:cells}) is composed of a sequence of cells, where each cell is a directed acyclic graph with nodes representing feature maps and edges representing network operations, e.g. convolutions or pooling layers (\cite{MANAS2019} and references therein). Given a set of possible operations, DARTS encodes the architecture search space with continuous parameters to form a one-shot model and performs searching by training the one-shot model with bi-level optimization, where the model weights and architecture parameters are optimized with training and validation data alternatively. MANAS uses a multi-agent learning approach (the \textit{search strategy}) to find the combination of operations leading to the best-performing architecture according the validation accuracy. 

\textbf{DARTS training configuration:} we follow the same experiment setting as in \cite{liu2018darts}. We replace the BN layers in DARTS with IN, LN, GN, PN and BGN in both search and evaluation stage. We search for 8 cells in $50$ epochs with batch size $64$ and initial number of channels as 16. We use SGD to optimize the model weights with initial learning rate $0.025$, momentum $0.9$ and weight decay $3\times10^{-4}$. Adam~\cite{kingma2014adam} is used to optimize architecture parameters with initial learning rate $3\times10^{-4}$, momentum $(0.5,0.999)$ and weight decay $10^{-3}$. We use network of 20 cells and 36 initial channels for evaluation to ensure a comparable model size as other baseline models. We use the whole training set to train the model for 600 epochs with batch size 96 to ensure convergence. For GN, we use $\rm G=32$ in \cite{wu2018group} while for BGN, we use $\rm G=256$ following Tab. \ref{Tab:ImageNet-batchsize}. Other hyper-parameters are set the same as the ones in the search stage. 

The best 20-cell architecture searched on CIFAR-10 by DARTS is trained from scratch with corresponding normalization methods used during the search phase. The validation accuracy of each method is reported in Tab.~\ref{tab:darts-nas}.
We can see that IN and LN fails to converge while BGN out-performs GN and PN significantly and outperforms BN slightly. The accuracy of BN is re-implication of \cite{liu2018darts}.
% It is worth noting that BN is used as the normalization layer in the neural architecture search phase, hence BN is at an advantage in this comparison.

\begin{table}[h]	
	\setlength{\tabcolsep}{0.07cm} 
	\caption{The validation accuracy on CIFAR-10 with using BN, IN, LN, GN, PN and BGN in DARTS for the search and evaluation phase.}
	\label{tab:darts-nas}
	\centering
	\begin{tabular}{c|ccccc|c}
		\hline %\toprule
		Normalization layer & BN & IN & LN & GN & PN & BGN \\
		\hline
	    accuracy & $97.33$ & - & - & $94.78$ & 94.41 & \textbf{97.40}  \\
		\hline
	\end{tabular}
\end{table}

\textbf{MANAS training configuration:} a single GPU is used to train the searched neural architectures (by BN) with replacing the normalization layers into BN, IN, LN, GN, PN and BGN. For GN, we use the best configuration $\rm G=32$ in \cite{wu2018group} while for BGN, we use $\rm G=64$. The network \textit{training protocol} is the same as in \cite{MANAS2019}, with the following hyperparameters: batch size $64$, epochs $600$, cutout length $16$, drop path probability $0.2$, gradient clip $5.0$, initial channels $36$, Cross-Entropy loss, SGD optimizer, learning rate decayed from $0.025$ to $0.0$, momentum $0.9$, weight decay $0.0003$.

The best 20-cell architecture searched on CIFAR-10 by MANAS is retrained from scratch with different normalization methods in place of the original BN used during the search phase. The validation accuracy of each method is reported in Tab.~\ref{tab:nas}.
We can see that IN, LN and PN fails to converge while BGN out-performs GN significantly and under-performs BN only slightly.
It is worth noting that BN is used as the normalization layer in the neural architecture search phase, hence BN is at an advantage in this comparison.

\begin{table}[h]	
	\setlength{\tabcolsep}{0.07cm} 
	\caption{The validation accuracy on CIFAR-10 with replacing the normalization layer in the nerual architecture searched by MANAS to BN, IN, LN, GN, PN and BGN.}
	\label{tab:nas}
	\centering
	\begin{tabular}{c|ccccc|c}
		\hline %\toprule
		Normalization layer & BN & IN & LN & GN & PN & BGN \\
		\hline
	    accuracy & \textbf{97.18} & - & - & $95.52$ & - & $97.15$  \\
		\hline
	\end{tabular}
\end{table}

DARTS experiment shows that BGN is generalizable to NAS for both search and evaluation. MANAS experiment shows that BGN is generalizable to less-regular neural architectures searched from NAS method.

\subsection{Adversarial Training on CIFAR-10}
\label{sec:adv}
DCNNs have been known to be vulnerable to malicious perturbed examples, known as adversarial attacks. Adversarial training was proposed to counter this problem. In this experiment, we apply BGN to adversarial training and compare its performance to BN, IN, LN, GN, and PN. 

\textbf{Implementation details:} the WideResNet \cite{BMVC2016_87} with the depth set as 10 and the wide factor set as 2 is used for image classification tasks on the CIFAR-10. The neural network is trained and evaluated against a four-step Projected Gradient Descent (PGD) attack. For the PGD attack, we set the step size as $2/255$, and the maximum perturbation norm as 0.0157. 200 epochs are trained until convergence. Due to the specialty of adversarial training, $\rm G=128$ is used in GN and BGN. It will divide images into patches, which can help to improve the robustness by breaking the correlation of adversarial attacks in different image blocks and constraining the adversarial attacks on the features within a limited range. This effect holds some similarity to the spectral normalization in \cite{farnia2018generalizable}. In the experiment, we use the Adam optimizer with a learning rate of 0.01. 

The robust and clean accuracy of training WideResNet with BN, IN, LN, GN, PN and BGN as the normalization layer are shown in Tab. \ref{tab:advresult}. The robust accuracy is more important than the clean accuracy in judging an adversarial network. PN experiences convergence difficulty and fails to converge. BGN out-performs BN and IN with a certain margin and out-performs LN and GN significantly.

\begin{table}[!ht]
	\setlength{\tabcolsep}{0.07cm} 
	\caption{The robust and clean validation accuracy of adversarial training with BN, IN, LN, GN, PN and BGN as the normalization layer in WideResNet. The clean/robust accuracy is evaluated on the clean/PGD attacked data.}
	\label{tab:advresult}
	\centering
	\begin{tabular}{c|ccccc|c}
		\hline
		Accuracy        &BN    &IN     &LN    &GN &PN    &BGN   \\ \hline 
		robust accuracy &48.79 &48.45  &44.15 &44.38 &$-$  &\textbf{49.64} \\\hline
	    clean accuracy  &72.09 &72.35  &64.1  &68.96 &$-$  &\textbf{72.94} \\\hline
	\end{tabular}

\end{table}

\subsection{Few Shot Learning}
\label{sec:few-shot}
\begin{figure*}[!t]
\centering
\includegraphics[width=0.80\textwidth]{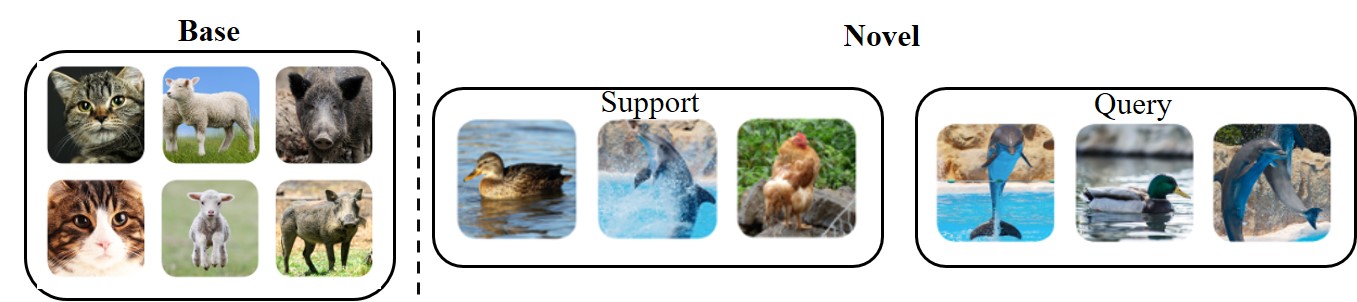}%
\caption{The illustration of FSL classification problem. The base category contains sufficient labelled samples, while the model aims at generalizing well on the novel category in the query set with limited labelled samples available in support set.}
\label{fig:FSL_intro}
\end{figure*}

\indent  We evaluate BGN on FSL task. FSL aims to train models capable of recognizing new, previously unseen categories using only limited training samples. Basically, a training dataset with sufficient annotated samples  comprise base categories. The test dataset contains $C$ novel classes, each of which is associated with only a few $K$ labelled samples (e.g. $\le 5$ samples) compose the support set, while  the remaining unlabelled samples consist the query set are used for evaluation (See Fig. \ref{fig:FSL_intro}). This is also referred to as a $C$-way $K$-shot FSL classification problem. 

\indent \textbf{Implementation details:} we experiment with imprinted weights \cite{qi2018low} model, which is one of the state-of-art metric-based FSL approaches and is widely used as the baseline in the current FSL community \cite{lifchitz2019dense,su2019boosting,gidaris2018dynamic}. 
At training time, a cosine classifier is learned on top of feature extraction layers and each column of classifier parameter weights can be regarded as a prototype for the respective class. At test time, a new class prototype (new column of classifier weight parameters) is defined by averaging the feature representation of support images, and the unlabelled images are classified via a nearest neighbor strategy. We test different settings, including 5-way 1-shot and 5-way 5-shot for the ResNet-12 backbone \cite{oreshkin2018tadam} on \emph{mini}ImageNet \cite{vinyals2016matching}.  We use the training protocol described in \cite{gidaris2018dynamic}: our model is optimized using SGD with Nesterov momentum set to $0.9$, weight decay to 0.0005, mini-batch size to 256, and 60 epochs. All input images were resized to $84\times84$. The learning rate was initialized to 0.1, and changed to 0.006, 0.0012, and 0.00024 at the 20th, 40th and 50th, respectively.

The mean accuracy of replacing the normalization layers in Imprinted Weights to BN, IN, LN, GN, PN and BGN, of training on \emph{mini}ImageNet, and of the 5-way 1-shot and 5-shot tasks are shown Tab. \ref{tab:fsl_method}. We can see that BGN out-performs BN slightly while out-performs IN, LN, GN and PN significantly, indicating the generalizability of BGN when the very limited labeled data is available.

\begin{table}[t]	
	\setlength{\tabcolsep}{0.07cm} 
	\caption{
	The mean accuracy of the 5-way 1-shot and 5-shot tasks on \emph{mini}ImageNet of Imprinted Weights with using ResNet-12 as a backbone. The normalization layer is replaced to BN, IN, LN, GN, PN and BGN. The mean accuracy of 600 randomly generated test episodes with 95\% confidence intervals is reported.}
	\label{tab:fsl_method}
	\centering
	\begin{tabular}{c|ccccc|c}
		\hline %\toprule
		%\multicolumn{2}{c}{Part}                   \\
		%\cmidrule(r){1-2}
% 		Model  &\multicolumn{2}{c}{\emph{mini}ImageNet} \\ 
		Model & BN & IN & LN & GN & PN & BGN\\ \hline
% 		DYNAMIC-FSL\cite{gidaris2018dynamic}  &60.19$\pm 0.84 $ &77.36  $\pm 0.61 $                            \\
1-shot &59.30 &52.17 &57.82 &56.55 &56.59 &\bf59.50  \\\hline
5-shot &76.22 &70.49 &74.87 &73.20 &73.89 &\bf76.32  \\\hline
	\end{tabular}
\end{table}

\subsection{Unsupervised Domain Adaptation on Office-31}

UDA aims to learn models on a target domain while annotations are only accessible in a related source domain. Normalization layers have effect of aligning feature distributions and reducing the domain gap~\cite{li2016revisiting, cariucci2017autodial}. We evaluate BGN and other normalization layers on a widely adopted UDA benchmark Office-31~\cite{saenko2010adapting}, which consists of 4110 images belonging to 31 classes, with three different domains: Amazon, Webcam and Digital SLR camera (DSLR). CAN~\cite{kang2019contrastive} is adopted as our model with replacing original BN with different normalization layers. 
% The results are summarized in Table~\ref{tab:office-uda}

% \textbf{Baseline method:} we adopt CAN~\cite{kang2019contrastive} as our base model, which is state-of-the-art UDA method on office-31 benchmark. It adapts model by aligning both domain's representation into a shared space and trains on source domain's data. Specifically, during the training, both source domain and target domain data are forwarded through a shared feature extractor to get relative representations. Then the target domain's labels are estimated by spherical K-means of extracted feature. After that, two domain's representations are aligned by simultaneously minimizing the intra-class MMD and maximizing the inter-class MMD, where MMD~\cite{gretton2007kernel} (\textbf{M}aximum \textbf{M}ean \textbf{D}iscrepancy) is a widely used metric for measuring two distribution's discrepancy. Aside from feature alignment, a traditional cross entropy loss is applied on source domain data to extract discriminative feature representation. When testing, the trained model is directly used for inferencing.

\textbf{Implementation details:} we follow the official released code's implementation of CAN and use ImageNet-pretrained ResNet-50 as the model's backbone. For tasks da (from domain DSLR to Amazon), wa and wd, the hyper-parameter $\rm G$ is set to 512. For ad and aw, we reduce $\rm G$ to 1 and 8 relatively as the source domain Amazon's backgrounds are totally white and may result in noisy statistics when the group size is small. For dw, $\rm G$ is set to 32.
We use Adam optimizer to optimize our model~\cite{kingma2014adam}. The learning rate is set to 0.001 and exponential learning rate decay is applied with decay rate 0.1 and decay step 25. The mini-batch size is set to 30. The training stops when the distance between source and target features' center is smaller than 0.001.

The results of BN, IN, LN, GN, PN and BGN are summarized in Table~\ref{tab:office-uda}. We can see that BGN outperforms other normalization layers in most adaptation tasks, especially in wa with an $1.5\%$ accuracy improvement.

\begin{table}[tbp]
  \centering
  \caption{The adaptation accuracy on Office-31 of CAN model with BN, IN, LN, GN, PN and BGN as the normalization layer. The result of each entry is averaged by three runs.}
    \begin{tabular}{l|ccccccc}
    \toprule
    model & ad & da & wa & aw & dw & wd & mean \\
    \midrule
    % BN(paper) & \multicolumn{1}{l}{95.0} & \multicolumn{1}{l}{78.0} & \multicolumn{1}{l}{77.0} & \multicolumn{1}{l}{94.5} & \multicolumn{1}{l}{99.1 } & \multicolumn{1}{l}{99.8} & 90.6 \\
    BN & 94.8 & 77.2  & 76.1  & 94.2 & 98.4  & 99.7  & 90.1  \\
    % G & 8 & 512 & 512 & 1 & 32 & 512\\
    BGN & \textbf{95.2}  & \textbf{78.5} & \textbf{78.5 } & 94.2 & \textbf{99.1}  & \textbf{99.9 } & \textbf{90.9}  \\
    GN & 90.0  & 77.2  & 77.2  & 91.1  & 96.9  & 99.1  & 88.6  \\
    IN & 88.0  & 76.1  & 75.4  & 90.0  & 97.2  & 98.5  & 87.5  \\
    LN & 92.8  & 76.8  & 76.0  & 91.1  & 97.8  & 98.7  & 88.9  \\
    PN & 90.9  & 76.5  & 77.1  & 90.6  & 97.8  & 99.5  & 88.7  \\
    \bottomrule
    \end{tabular}%
  \label{tab:office-uda}%
\end{table}%

\section{Conclusion}
BGN is proposed with good performance, stability and generalizability and without using additional trainable parameters, information across multiple layers or iterations, or extra computation. BGN facilitates the noisy/confused statistic calculation in BN with adaptively introducing feature instances from the grouped (channel, height and width) dimensions and uses a hyper-parameter $\rm G$ to control the size of divided feature groups. It is intuitive to implement, is orthogonal to and can be used in addition to many methods reviewed in Related Work to further improve performance.

\clearpage

\bibstyle{aaai21.bst}
\bibliography{AAAI2021.bib}

\end{document}